\pdfoutput=1

\documentclass[11pt]{article}

\usepackage[final]{acl}
\usepackage{times}
\usepackage{latexsym}
\usepackage[T1]{fontenc}
\usepackage[utf8]{inputenc}
\usepackage{microtype}
\usepackage{inconsolata}
\usepackage{graphicx}
\usepackage{enumitem}
\usepackage{adjustbox}
\usepackage{tikz}
\usetikzlibrary{positioning, fit, decorations.pathreplacing}
\usepackage{mathtools}
\usepackage{amsmath}
\usepackage{amsfonts}
\usepackage{amssymb}
\usepackage{bbm}
\usepackage{tabularx}
\usepackage{makecell}
\usepackage{xcolor}
\usepackage{cleveref}
\usepackage{booktabs}
\usepackage{hyperref}
\usepackage{multirow}

\definecolor{hl1}{RGB}{210, 20, 20}
\definecolor{hl2}{RGB}{20, 20, 200}
\definecolor{hl3}{RGB}{0, 160, 0}

\usepackage{framed}
\usepackage{CJKutf8}
\usepackage[most]{tcolorbox}
\tcbuselibrary{breakable,skins}
\tcbset{colback=white,colframe=black,breakable,size=fbox,enhanced jigsaw}

\newenvironment{errorbox}{}{}
\tcolorboxenvironment{errorbox}{
  breakable,
  enhanced,
  colback=gray!5,
  colframe=gray!50,
  coltitle=black,
  fonttitle=\bfseries,
  title={},
  sharp corners=south,
  boxrule=0.4pt, 
  right=2pt,
  top=2pt,    
  bottom=2pt,
  before skip=6pt,
  after skip=8pt,
  boxsep=2pt           
}

\usepackage{listings}

\title{Classifying and Addressing the Diversity of Errors in\\ Retrieval-Augmented Generation Systems}

\author{Kin Kwan Leung \ \ \ Mouloud Belbahri  \ \ \ Yi Sui \ \ \ Alex Labach \\ {\bf Xueying Zhang \ \ \ Stephen Anthony Rose \ \ \ Jesse C. Cresswell} \\
        Layer 6 AI, Toronto, Canada
        \\
    \small{
    \texttt{\{kk, mouloud, amy, alex, chloe, stephen, jesse\}@layer6.ai}
    }
    }

\begin{document}
\maketitle
\begin{abstract}

Retrieval-augmented generation (RAG) is a prevalent approach for building LLM-based question-answering systems that can take advantage of external knowledge databases. Due to the complexity of real-world RAG systems, there are many potential causes for erroneous outputs. Understanding the range of errors that can occur in practice is crucial for robust deployment. We present a new taxonomy of the error types that can occur in realistic RAG systems, examples of each, and practical advice for addressing them. Additionally, we curate a dataset of erroneous RAG responses annotated by error types. We then propose an auto-evaluation method aligned with our taxonomy that can be used in practice to track and address errors during development. Code and data are available at \url{https://github.com/layer6ai-labs/rag-error-classification}.
\end{abstract}

\section{Introduction}\label{sec:intro}

Retrieval-augmented generation (RAG) \cite{lewis2020rag} has become the dominant paradigm for applying generative large language models (LLMs) in applications where outputs must incorporate knowledge from outside of the model's training set. This is especially valuable for grounding generation in factual information to reduce fabricated content \cite{maynez2020faithfulness, shuster2021retrieval}, and when non-public domain knowledge is required. RAG systems are already widely deployed in real-world applications to provide natural language interfaces to knowledge sources \cite{amugongo2024}, but despite their merits they can still be error-prone in practice~\cite{venkit2024search,magesh2024hallucination,grant2024google}. Due to the greater complexity of RAG pipelines compared to direct LLM generation, these errors are diverse and their causes can be difficult to trace. Deploying a RAG pipeline, especially in critical industries like healthcare, requires understanding the variety of errors that can occur in order to monitor and minimize them.

Existing work on RAG errors has generally not accounted for the complexity of real-world RAG systems and their failure modes. On the data side, widely used benchmark tasks for evaluating RAG systems are often overly simplistic, typically featuring multiple-choice questions \cite{mihaylov2018suit, guinet2024autometed, yang2024crag} or requiring short factual answers for ease of validation (e.g., \citet{rajpurkar2016squad, yang2018hotpotqa, joshi2017triviaqa, kwiatkowski2019NQ}). In practice, users expect more sophisticated answers that thoroughly explain a topic, but as a result can fail in more subtle ways. 

\begin{figure*}[t]
\centering
\input{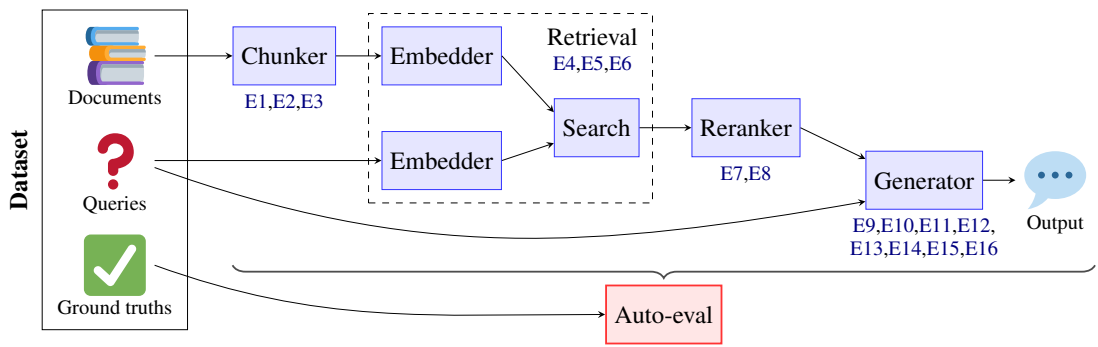}
\caption{Diagram of our implemented RAG architecture, reflective of systems currently applied in industry. Components are annotated with error types that are caused by them.}
\label{fig:rag_pipeline}
\end{figure*}

On the model side, real RAG systems use complex multi-step pipelines that go beyond a retriever-generator pair~\cite{akkiraju2024facts}, with additional data processing steps like adaptive chunking and reranking. Prior works that categorize RAG errors have used bare-bones pipelines benchmarked on simple datasets, and as a result have overlooked entire classes of errors, especially those associated with pre-generation steps \citep{barnett2024seven, venkit2024search}. As a result, practitioners relying on these works may be left with an overly optimistic view of their RAG system's performance, and fail to identify the scope and severity of errors afflicting them. To address these gaps, we use challenging public datasets and build a realistic RAG question-answering (QA) system reflective of those currently used in industry. We perform a deep analysis of RAG failure modes with illustrative examples, and practical advice for mitigating errors.

Finally, we develop auto-evaluation tools to classify error types according to our taxonomy. To validate these systems, we manually annotate errors producing a first-of-its-kind RAG error type dataset. The overall purpose of our taxonomy and auto-evaluation system is for practitioners to be able to identify weak links and common errors in their RAG pipelines. Hence, we provide an end-to-end demonstration of how auto-evaluation can be used by identifying the most common errors produced by our reference pipeline, and implementing targeted improvements.

Our main contributions are:
\begin{enumerate}[itemsep=0pt, topsep=0pt, parsep=0pt, partopsep=0pt]
\setlength{\leftskip}{-2pt}
	\item A novel taxonomy of errors made by RAG systems, along with examples of each and recommendations for addressing them;
    \item An auto-evaluation system for identifying and classifying errors according to our taxonomy;
    \item A dataset of RAG errors annotated by type.
\end{enumerate}

\section{Related Works}\label{sec:related}

\textbf{RAG Error Taxonomies:} \citet{barnett2024seven} provide the most comparable taxonomy of RAG errors to ours with seven types associated to select RAG components. Here, we describe a more comprehensive range of RAG errors relating to \emph{every} step of a realistic RAG pipeline, along with specific examples and mitigation strategies, both of which are lacking in the prior work. \citet{venkit2024search} classify errors in public black-box QA systems from the perspectives of human evaluators without examining specific pipeline components or attributing errors to them. \citet{Agrawal2024Mindful} focus on RAG systems built on knowledge graphs. \citet{yu2024evaluation} survey RAG evaluation benchmarks and metrics but do not break down causes of errors beyond the high-level dichotomy of retrieval versus generation. Other works \cite{huang2025survey, magesh2024hallucination} that specifically focus on hallucinations overlook the nuanced ways RAG can fail. Error taxonomies for non-RAG QA settings also exist~\cite{rawte2023troubling, huang2025survey}.

\textbf{RAG Auto-evaluation:} Simple QA benchmarks can be evaluated using exact match or overlap-based metrics, but realistic questions and answers require more sophisticated evaluations based in natural language understanding. Previous works have applied metrics using LLM-as-a-Judge \cite{zheng2023llmasajudge} to automatically evaluate RAG outputs, including RAGChecker~\cite{ru2024ragchecker}, ARES~\cite{saad-falcon2024}, and RAGAs~\cite{es2024ragas}. These works are each tailored to a specific error classification scheme, with narrower scope than ours. Other prior works have developed benchmark datasets with LLM-based auto-evaluation (e.g., \citet{zhu2024rageval, liu2024cofe}), but do not probe the internals of RAG systems to give detailed information on error causes.

\section{RAG Implementation}\label{sec:implementation}

To demonstrate errors that can occur in realistic, production-grade RAG systems, we designed a modular pipeline reflecting common architectural patterns used in industry \cite{nvidia2024ragreranking, microsoft2025advancedrag} (see \Cref{fig:rag_pipeline}). We begin with a \textbf{chunking} stage, where documents are processed to optimize granularity and relevance. The chunks are indexed using an embedding model, then queried via  dense \textbf{retrieval}, which returns top candidates based on similarity to the input query. Retrieved chunks are \textbf{reranked} using a separate language model given the original query as context, which can better highlight semantic relevance. Finally, the top-ranked chunks are fed to a \textbf{generator} LLM, which produces the final answer. This architecture supports component-level variation and allows for in-depth error analysis at each stage of the pipeline. We provide complete details on the options implemented and used at each stage in \Cref{app:a}.

\textbf{Datasets:} We use the DragonBall dataset, part of the RAGEval framework \citep{zhu2024rageval}, an evaluation suite designed to rigorously assess RAG systems across diverse domains and scenarios. The dataset provides a broad benchmark for evaluating RAG performance in complex settings, spanning domains such as finance, law, and medicine. The dataset also comes with query type annotations in seven categories, which allows us to perform thorough analysis on error classification. We used both the English and Chinese portions of the dataset with 3108 and 3601 questions respectively. We also use the CLAPnq dataset for additional analysis~\citep{rosenthal2025clapnq}, which contains high-quality reference answers and passages for 4946 natural questions relating to Wikipedia documents.

\vspace{-4pt}
\section{Error Classification}\label{sec:classification}
\vspace{-4pt}

We present a practically grounded taxonomy of errors that can be attributed to different parts of a RAG system. Errors are grouped by the pipeline stage that caused them: chunking, retrieval, reranking, or generation. For each stage and error type, we describe the nature of the error, explain why it occurs in RAG systems, provide real examples from our RAG implementation, and finally give advice for how to reduce their occurrence.

Error types are not mutually exclusive; in practice multiple error types often co-occur, as errors early in the pipeline beget later ones. Additionally, we do not claim that our taxonomy is exhaustive, since variations in RAG architecture will lead to different error types. Our taxonomy focuses on errors that occur within RAG systems, and hence we exclude failures caused by adversarial inputs, faults in the corpus, or similar anomalies.

\vspace{-4pt}
\subsection{Chunking}\label{sec:chunking}
\textbf{E1 Overchunking:}\phantomsection\label{e:1} Documents are split into excessively small or disjointed segments, causing incomplete coverage of topics. Individual chunks are fragmented or ambiguous. Errors cascade downstream when search fails to retrieve consecutive chunks.

\begin{errorbox}    \noindent\textbf{Query}: What platform did Sunrise Holidays introduce in April 2019? \\
    \textbf{Chunk Excerpt}: \textit{[Excluded from chunk: In April 2019, Sunrise Holidays introduced an \textcolor{hl1}{online booking platform}, which greatly improved its competitiveness.]} The launch of this user-friendly platform attracted more customers [...]  \\
    \looseness=-1\textbf{Response}: The platform introduced by Sunrise Holidays in April 2019 is not specified.\\
    \textbf{Ground Truth}: \textcolor{hl1}{An online booking platform.} \\
    \textbf{Cause}: The retrieved chunks come from the correct document, but the description of it as an \textit{online booking} platform is cut off.
\end{errorbox}

\noindent
\textbf{E2 Underchunking:}\phantomsection\label{e:2} Chunks are too large, covering multiple topics with mixed content. Irrelevant information dilutes keywords or phrases, lowering retrieval scores on the correct chunks. Chunks provided to the generator contain extraneous content that can confuse the model.

\begin{errorbox}
	\noindent\textbf{Query}: What system does CleanCo Housekeeping Services have in place to safeguard assets and ensure financial accuracy? \\
    \textbf{Chunk Excerpt}: [...] To mitigate risks such as increasing competition, regulatory changes, and economic uncertainties, CleanCo plans to implement \textcolor{hl1}{risk management} strategies through diversification and continuous monitoring. [...] \\
	\textbf{Response}: CleanCo has a system in place to safeguard assets and ensure financial accuracy through its \textcolor{hl1}{risk management} framework. \\
	\textbf{Ground Truth}: Unable to answer. \\
	\textbf{Cause}: Retrieved chunks contain so much tangential information that the generator uses unrelated information to answer instead of abstaining.
\end{errorbox}

\noindent
\textbf{E3 Context Mismatch:}\phantomsection\label{e:3} Chunks split text at arbitrary points, breaking contextual links by separating definitions from the information they support. This ambiguity causes failed retrieval downstream where keywords are missing.

\begin{errorbox}
    \looseness=-1\textbf{Query}: Why is \textcolor{hl1}{the Philippines} typhoon prone? \\
    \textbf{Chunk A (Retrieved)}: \textcolor{hl1}{The Philippines'} evident risk to natural disasters is due to its location. [...] \\
    \textbf{Chunk B (Not Retrieved)}: In addition, \textcolor{hl2}{the country} faces the Pacific Ocean where 60\% of the world's typhoons are made. [...] \\
	\textbf{Response}: Due to its geographical location, climate, and topography. \\
    \textbf{Ground Truth}: It faces the Pacific Ocean where 60\% of the world's typhoons are made. \\
	\textbf{Cause}: Chunk B mentions ``\textcolor{hl2}{the country}'' 
 but not \textcolor{hl1}{the Philippines} by name, which led to the retriever incorrectly assigning it low relevance.
\end{errorbox}

\noindent
\textbf{Improvement}:
Straightforward heuristics include adjusting chunk size, where larger chunks help reduce \hyperref[e:1]{E1}, smaller chunks help reduce \hyperref[e:2]{E2}, while adding small overlaps mitigates \hyperref[e:3]{E3} by preserving continuity~\cite{safjan2023chunking}. However, these approaches require careful tuning and are often insufficient on their own. Adaptive chunking strategies handle real-world variability more effectively.

Structure-aware chunking preserves document logic by splitting along units like paragraphs or section headers. This works especially well for corpora with consistent structure, such as financial reports~\cite{yepes2024financial}. Semantic chunking segments text based on meaning, such as by detecting topic shifts using cosine distance between sentence embeddings~\cite{qu2025semantic} or spikes in perplexity~\cite{zhao2024meta}. Hybrid strategies combine structural and semantic cues to balance coherence and topical focus. For example, S2 Chunking~\cite{verma2025s2chunkinghybridframework} integrates layout with embeddings and achieves strong results on document QA.

\subsection{Retrieval}
\textbf{E4 Missed Retrieval:}\phantomsection\label{e:4} Relevant chunks are not retrieved, leading the generator to give incomplete answers, fabricate information to fill gaps, or abstain unnecessarily. Misses can occur when high-similarity chunks only address part of the question.

\begin{errorbox}
	\noindent\textbf{Query}: What were the large-scale financing activities undertaken by \textcolor{hl1}{Vanguard Media Group} in 2019 and \textcolor{hl2}{Energex Corp.} in 2021? \\
	\textbf{Response}: \textcolor{hl1}{Vanguard Media Group}'s large-scale financing activities in 2019 included the acquisition of Silver Screen Studios for \$200 million [...] \\
	\textbf{Ground Truth}: \textcolor{hl1}{Vanguard Media Group} raised \$100 million in capital in 2019, while \textcolor{hl2}{Energex Corp.} conducted large-scale financing in 2021. \\
	\textbf{Cause}: The query asks about two entities, but all top-ranked chunks relate to only one. 
\end{errorbox}

\noindent
\textbf{E5 Low Relevance:}\phantomsection\label{e:5} Retrieved chunks are only loosely related to the query. Causes include an unbalanced corpus with many chunks relating to one topic, and vocabulary mismatch where the query uses different terms than the corpus. 

\begin{errorbox}
	\noindent\textbf{Query}: In which year is the Corporate Governance Report for Green Fields Agriculture Co.? \\
	\textbf{Response}: The Corporate Governance Report does not provide a specific year, but it appears to cover events and initiatives from \textcolor{hl1}{2018} \textcolor{hl2}{and 2021}. \\
	\textbf{Ground Truth}: \textcolor{hl1}{2018}. \\
	\textbf{Cause}: Many chunks from the Report were assigned high similarity and retrieved, but did not relate to the Report's publication year. The generator latches onto years discussed in the content.
\end{errorbox}

\noindent
\textbf{E6 Semantic Drift:}\phantomsection\label{e:6} Retrieved documents match keywords, not the query's intent, due to reliance on keyword matching rather than semantic relevance.

\begin{errorbox}
	\noindent\textbf{Query}: According to the judgment of Hartford, Ashland, Court and Glenwood, Quailwood, Court, what sentences were given to the defendants \textcolor{hl1}{M. Cooper} and \textcolor{hl2}{Y. Nelson}? \\
	\textbf{Response}: \textcolor{hl1}{M. Cooper} was sentenced to four years imprisonment. \\
	\textbf{Ground Truth}: \textcolor{hl1}{M. Cooper} was sentenced to four years imprisonment; \textcolor{hl2}{Y. Nelson} was sentenced to three years imprisonment. \\
	\textbf{Cause}: The retrieved chunks contain information about both defendants' cases, but don't include the sentence for \textcolor{hl2}{Y. Nelson}.
\end{errorbox}

\noindent
\textbf{Improvements:} Retrieval systems present the widest diversity of approaches out of the RAG stages, and hence can be tuned in many ways.
First, query rewriting via keyword expansion, paraphrasing, or semantic reformulation can improve recall and precision by 
reducing vocabulary mismatch~\cite{ma2023query}. Second, hybrid retrieval that combines sparse (e.g., BM25 \cite{bm25}) and dense (embedding-based) methods can improve robustness by capturing both lexical and semantic matches~\cite{ni2022large}. Third, top-$k$ retrieval is often insufficient when both \hyperref[e:4]{E4} and \hyperref[e:5]{E5} are frequent. Adaptive thresholds can better balance recall and precision~\cite{sun2025dynamic}. Fourth, domain-specific or fine-tuned embedding models can significantly boost retrieval accuracy and reduce hallucinations~\cite{databricks2025retrieval}. Finally, metadata filtering enhances precision by using structured fields (e.g., source or section) to prioritize relevant content~\cite{Poliakov2025multi}. 

\vspace{-6pt}
\noindent
\subsection{Reranking}
\vspace{-2pt}
\textbf{E7 Low Recall:}\phantomsection\label{e:7} Although the necessary chunks are retrieved, they are reranked too low and not forwarded to the generation model. Low recall produces false negatives, where relevant data is available but unintentionally ignored.

 \begin{errorbox}
 	\noindent\textbf{Query}: According to Greenfield, Fairview, Court and Vandalia, Bayside, Court, whose sentencing time is longer, H. Price or G. Torres? \\
 	\textbf{Response}: G. Torres' sentencing time is longer, four years, compared to \textcolor{hl2}{H. Price, whose sentence is not mentioned in the provided context}. \\
 	\textbf{Ground Truth}: \textcolor{hl1}{H. Price was sentenced to eight years of imprisonment}, while G. Torres was sentenced to four years imprisonment. \\
 \textbf{Cause}: The chunk with H. Price's sentencing was retrieved, but was ranked below the reranker threshold and not passed to the generator.
 \end{errorbox}

\noindent
\textbf{E8 Low Precision:}\phantomsection\label{e:8} Irrelevant chunks are ranked highly and forwarded to the generation model, leading to the generation model being confused by noise. Low precision leads to false positives, where the generator is overly confident in noisy context.

\begin{errorbox}
	\noindent\textbf{Query}: On which stock exchange is Retail Emporium publicly traded? \\
	\textbf{Response}: The \textcolor{hl1}{New York} Stock Exchange. \\
	\textbf{Ground Truth}: Unable to answer. \\
	\textbf{Cause}: Highly ranked but irrelevant chunks reference \textcolor{hl1}{New York}, leading the model to make an inappropriate conclusion.
\end{errorbox}

\noindent
\textbf{Improvements:} 
Modern RAG pipelines increasingly include a reranking stage to reorder retrieved chunks based on semantic relevance \cite{nvidia2024ragreranking}. The two dominant approaches are cross-encoders, which jointly encode query-chunk pairs~\cite{déjean2024thorough}, and LLM-based rerankers, which use prompted models to assess candidates~\cite{sun2023chatgpt}. Rerankers may still miss key context \hyperref[e:7]{(E7)} or over-rank irrelevant chunks \hyperref[e:8]{(E8)} when domain relevance is poorly understood. Fine-tuning on in-domain data is the most effective way to improve performance~\cite{moreira2024enhancing, aarsen2025trainreranker, yu2024rankrag}, and even small amounts of labeled data can yield significant gains~\cite{aarsen2025trainreranker}. More specifically, incorporating hard negatives, which are plausible but non-relevant passages, further helps the model reduce false positives \hyperref[e:8]{(E8)}~\cite{yu2024rankrag}.

\subsection{Generation}
\textbf{E9 Abstention Failure:}\phantomsection\label{e:9} The model should have abstained, but instead answers incorrectly. Abstention is appropriate when the answer is not available in the corpus, or when the query is ambiguous.

\begin{errorbox}
	\noindent\textbf{Query}: How much did Green Fields Co. raise through large-scale financing in December 2018? \\
	\textbf{Response}: Green Fields Co. did not raise any funds in December 2018. \\
	\textbf{Ground Truth}: Unable to answer \\
	\textbf{Cause}: The retrieved context contains information about 2018, but not about fundraising. Hence, the model assumes fundraising did not occur, but this was not stated in the corpus.
\end{errorbox}

\noindent
\textbf{E10 Fabricated Content:}\phantomsection\label{e:10} Although the query is answerable, the response includes unverifiable information not grounded in the retrieved chunks nor supported by external evidence.

\begin{errorbox}
	\noindent\textbf{Query}: According to the hospitalization records of Southport General Hospital, summarize the past medical history of K. Mendoza. \\
	\textbf{Response}: K. Mendoza has a history of \textcolor{hl2}{Parkinson's disease}, with symptoms including tremors, bradykinesia, and postural instability. \\
	\textbf{Ground Truth}: K. Mendoza was \textcolor{hl1}{generally healthy} prior to the onset of his symptoms. \\
	\textbf{Cause}: The model fabricates a history of \textcolor{hl2}{Parkinson's disease} which is not in the context.
\end{errorbox}

\noindent
\textbf{E11 Parametric Overreliance:}\phantomsection\label{e:11} The LLM relies on its internal (parametric) knowledge rather than retrieved documents. Internal knowledge may be factual, as opposed to fabricated, but irrelevant.

\begin{errorbox}
	\noindent\textbf{Query}: What was the total amount of liabilities of Artistic Creations at the end of 2019? \\
	\textbf{Response}: The total amount of liabilities of Artistic Creations at the end of 2019 is \$150M - \$20M (acquisition) - \$50M (financing) - \$10M (investment) + \$15M (restructuring) = \$85M. \\
	\textbf{Ground Truth}: Unable to answer. \\
	\textbf{Cause}: The model does a calculation based on parametric financial knowledge instead of acknowledging that it has insufficient information.
\end{errorbox}

\noindent
\textbf{E12 Incomplete Answer:}\phantomsection\label{e:12} The response is from the corpus and correct, but misses critical details.

\begin{errorbox}
	\noindent\textbf{Query}: How did the dividend in July 2017 reflect AMS Inc.'s \textcolor{hl2}{strong financial performance}? \\
	\textbf{Response}: The dividend in July 2017 reflected AMS Inc.'s \textcolor{hl2}{strong financial performance}, which was driven by market demand and changes in product prices, resulting in an operating income of \$450 million and a net profit of \$60 million. \\
	\textbf{Ground Truth}: The dividend demonstrated the company's \textcolor{hl1}{ability to generate sufficient profits and maintain a healthy cash flow}. \\
	\textbf{Cause}: The response affirms the basis of the question, but does not analyze the situation.
\end{errorbox}

\noindent
\textbf{E13 Misinterpretation:}\phantomsection\label{e:13} The generator misuses or misrepresents retrieved content.

\begin{errorbox}
\textbf{Query}: According to the judgments of Trenton, Springfield and Hamilton, Harrison Courts what sentences were given to \textcolor{hl1}{Q. Ruiz} and \textcolor{hl2}{M.~Ward}? \\
	\textbf{Response}: \textcolor{hl2}{M. Ward} was sentenced to four years imprisonment, and \textcolor{hl1}{Q. Ruiz} has the right to appeal this judgment within 30 days. \\
	\textbf{Ground Truth}: \textcolor{hl2}{M. Ward} was sentenced to four years of imprisonment, and \textcolor{hl1}{Q. Ruiz} was sentenced to 5 years of imprisonment.\\
	\textbf{Cause}: The model misinterprets facts about two defendants as being related and incorrectly combines them for the answer.
\end{errorbox}

\noindent
\textbf{E14 Contextual Misalignment:}\phantomsection\label{e:14} The response is factual and comes from related information in the corpus, but does not address the query.

\begin{errorbox}
	\noindent\textbf{Query}: According to the judgment of Preston, Lancaster, Court, summarize the \textcolor{hl1}{facts of M. Harris's crimes}. \\
	\textbf{Response}: M. Harris was \textcolor{hl2}{found guilty} of homicide and was \textcolor{hl2}{sentenced} to life imprisonment. \\
	\textbf{Ground Truth}: M. Harris committed the crime of intentional homicide by \textcolor{hl1}{tampering with the brakes} of her business partner. \\
	\textbf{Cause}: The response contains factual information, but discusses \textcolor{hl2}{sentences} rather than the requested \textcolor{hl1}{facts of the crimes}.
\end{errorbox}

\noindent
\textbf{E15 Chronological Inconsistency:}\phantomsection\label{e:15} The response presents events or facts in the wrong temporal order, or confuses past, present, or future timelines.

\begin{errorbox}
	\noindent\textbf{Query}: 
    Which company, Green Fields or JetWing, changed their board of directors earlier? \\
	\textbf{Response}: Green Fields made changes to their board of directors earlier, in \textcolor{hl1}{December 2021}, while JetWing made changes in \textcolor{hl2}{October 2021}. \\
	\textbf{Ground Truth}: JetWing made changes earlier in \textcolor{hl2}{September 2021}, while Green Fields made changes in \textcolor{hl1}{December 2021}. \\
	\textbf{Cause}: The model provides a wrong date, and incorrectly considers Dec. earlier than Oct.
\end{errorbox}

\noindent
\textbf{E16 Numerical Error:}\phantomsection\label{e:16} The response includes incorrect calculations, quantities, or misrepresents numerical data from the retrieved documents.

\begin{errorbox}
	\noindent\textbf{Query}: 
	Which company reduced its liabilities more, InnovateTech or Sparkling Clean? \\
    \textbf{Response}: InnovateTech reduced its liabilities by a larger amount, a \textcolor{hl1}{total of \$50 million}, compared to Sparkling Clean's £500,000. \\
	\textbf{Ground Truth}: InnovateTech reduced its liabilities by a larger amount (\textcolor{hl2}{\$30 million}) compared to Sparkling Clean (£500,000).\\
	\textbf{Cause}: The model incorrectly added together two numbers from the context giving \$50 million.
\end{errorbox}

\noindent
\textbf{Improvements:} 
Generation-stage errors can be mitigated through abstention modeling, answer validation, prompt engineering, and tool augmentation. 
For abstention failures \hyperref[e:9]{(E9)}, models can explicitly use ambiguity detection~ \cite{ambiguousquestions, kim2023tree}, conformal abstention \cite{conformalabstention}, or fine-tuning approaches like Trust-Align~\cite{songmeasuring}. For content fabrication \hyperref[e:10]{(E10)},  post-generation validation such as Chain-of-Verification~\cite{dhuliawala2024chain} or critique modules~\cite{asai2023self} encourage fact-checking and grounding. Errors related to context misuse (\hyperref[e:11]{E11}-\hyperref[e:14]{E14}) often stem from noisy retrieval; refinement modules can filter context before generation~\cite{contextrefinement, chirkova2025provence, wang2025speculative}. Query decomposition~\cite{lin2023decomposing}  helps reduce \hyperref[e:12]{E12} by breaking complex queries into sub-questions, especially for comparisons, causal questions, or multi-step reasoning.
For temporal and numerical errors (\hyperref[e:15]{E15}–\hyperref[e:16]{E16}), structured prompting~\cite{wei2022chain}, timeline reasoning~\cite{bazaga2025learningreasontimetimeline}, or tool-augmented generation with code modules like PAL~\cite{gao2023pal} can boost reliability.

\section{RAGEC: RAG Error Classification}\label{sec:auto-eval}

To deepen our understanding of the RAG error taxonomy and aid practitioners in applying it, we developed a RAG Error Classification system (RAGEC). In this section we describe its design, curate a dataset with human annotations to validate it, and demonstrate its use via our reference RAG system with standard research datasets.

The primary objective of RAGEC is to identify weak links in the RAG pipeline by classifying the \emph{stage} responsible for errors. More granular error \emph{types}, which can co-occur and overlap, are classified secondarily. Identifying the \emph{first} stage where an error occurred enables the developer to implement targeted improvements on the RAG pipeline.

\textbf{Design.} Given a RAG pipeline and dataset, RAGEC consists of 3 steps: answer evaluation, stage classification, and error type classification. For \textit{answer evaluation}, we prompt an LLM on each datapoint individually to determine whether the generated answer is incorrect. Included as context are the original query, ground-truth answer and documents according to the dataset, and the generated answer. This evaluation yields a subset of $N_{\mathrm{err}}$ examples identified as incorrect. For \textit{stage classification} on an erroneous example, we apply a rules-based approach to cascade over the stages, using specialized information to determine if an error is present. Starting with the generation stage, we check whether the generator had sufficient information to answer the query by computing the chunk-level recall and comparing to a threshold. If so, we conclude the generation stage caused the error. Otherwise, we continue and check if the reranker recall dropped compared to the retrieval recall, indicating that the reranker caused the error. If not, we move to the final comparison between retrieval and chunking. To differentiate these two we adapt the idea from \citet{walk-the-talk} to extract key concepts in the query. We then prompt the LLM as to whether each concept is included in the ground truth chunks. If query concepts fail to appear in the chunks, it indicates the chunking stage caused the error. Otherwise, by process of elimination, we conclude the retriever was responsible. This design is empirically motivated, and provided the highest alignment with our human-annotated results. Other designs we tested are described in \Cref{app:alternatives}.

\begin{table}[t]
\centering
\small
\caption{Stage classification agreement matrix}
\label{tab:human_annotation_stage}
\setlength{\tabcolsep}{2pt} 
\begin{tabular}{lcccc}
\toprule
\textbf{Human}&\multicolumn{4}{c}{\textbf{RAGEC Stage Classification}}\\
\textbf{Annotation} & \textbf{Chunking} & \textbf{Retrieval} & \textbf{Reranking} & \textbf{Generation} \\
\midrule
\textbf{Chunking} & 49 & 13 & 5 & 10 \\
\textbf{Retrieval} & 24 & 85 & 12 & 40\\ 
\textbf{Reranking} &6 & 8 & 22 & 10\\
\textbf{Generation} & 4 & 21 & 6 & 62\\
\bottomrule
\end{tabular}
\end{table}

For \textit{error type classification}, we take information relevant to the identified stage, and conduct detailed analysis using LLM-as-a-Judge \citep{zheng2023llmasajudge}. For example, on a chunking error we only provide the query, ground truth answer, and chunks from the ground truth document. Other information, like the retrieved chunks or generated answer are irrelevant for classifying a chunking error type. To account for uncertainty in the classification, each erroneous example is annotated $K$ times \citep{selfconsistency}. We then compute the modal vote over error types and metrics that capture the self-consistency of the LLM's predictions.

Further details on the implementation of RAGEC, including the complete methodologies, inputs used for each step, and LLM prompts, are included in \Cref{app:b}.

\textbf{Data Curation.} No existing dataset details the types of errors which can occur in RAG systems. Hence, to validate RAGEC, we manually annotated 406 erroneous responses generated by our reference system on the DragonBall dataset as to the responsible stage, and error types present. Annotators used all available context including ground truth answer, generated response, retrieved and reranked chunks, and the corpus. First, annotators selected the earliest stage in the pipeline which demonstrated erroneous behaviour, since errors propagate through RAG systems. Subsequently, one or more error types were selected within that stage. The repo at \url{https://github.com/layer6ai-labs/rag-error-classification} contains our annotated data.

\begin{figure}[t]
  \centering
  \includegraphics[width=0.9\columnwidth]{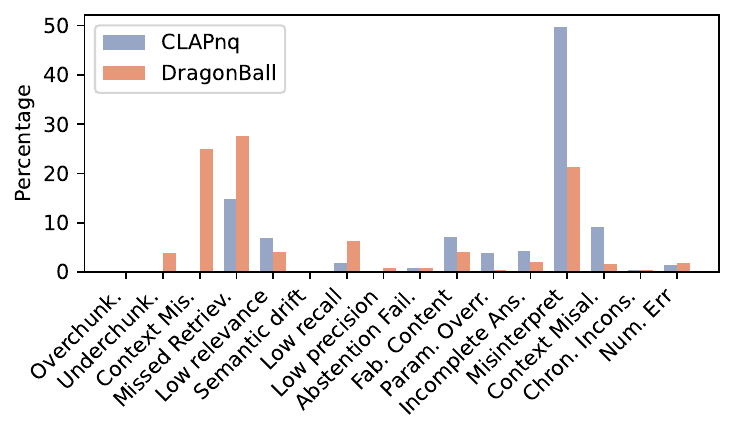}
  \caption{Modal error type distribution per dataset. 
  }
  \label{fig:mode_error_counts_per_dataset}
\end{figure}

\textbf{Results \& Analysis.} In the first step of RAGEC, \textit{answer evaluation}, $27\%$ ($25\%$) of responses for the DragonBall-EN (CLAPnq) dataset were classified as incorrect, giving $N_{\mathrm{err}}=832\ (1222)$ responses to analyze. Of these, 406 queries were manually annotated, and 377 were confirmed as erroneous, reaching an agreement rate of 92.9\%.

For \textit{stage classification}, RAGEC achieves a 57.8\% agreement rate with human annotations. Details of the results are shown in \Cref{tab:human_annotation_stage}. We find that human annotations more often identify the retrieval stage as causing errors, whereas RAGEC favors blaming the generator. While RAGEC's agreement rate is not close to perfect, we tested several other methods to engineer the context for LLM-based auto-evaluation that all performed worse (described in \Cref{app:alternatives}). Hence, RAG error classification appears to be a challenging problem due to the highly complex set of intermediate outputs produced by RAG pipelines. We encourage the community to further explore this topic by releasing our reference pipeline and annotated error data.

For \textit{error type classification}, we query the LLM $K=10$ times for each datapoint and extract the most and second-most frequently predicted error types. Compared to human-annotated error types, RAGEC achieves an accuracy of 40.3\%. \Cref{fig:mode_error_counts_per_dataset} shows the distribution of the modal error type for DragonBall-EN and CLAPnq. \Cref{tab:mode_error_counts} shows more details for the DragonBall-EN dataset (CLAPnq is shown in \Cref{tab:mode_error_counts_clapnq} of  \Cref{app:clapnq-distribution}). 
These distributions differ, showing that the nature of the dataset can affect the types of errors committed by a given RAG pipeline.
This type of information can direct diagnosis and triaging of problems in RAG. For example, CLAPnq exhibited no chunking errors, as its source documents were manually curated by the dataset annotators into self-contained  paragraphs designed to answer the corresponding queries. In contrast, DragonBall-EN almost never shows Parametric Overreliance \hyperref[e:11]{(E11)} for both English and Chinese (See Appendix~\ref{app:chinese-dragonball} for details). This can be attributed to the dataset’s domain-specific and context-dependent queries, which are unlikely to be covered by the LLM’s parametric knowledge.
RAGEC also demonstrates that some error types dominate their stage, like \hyperref[e:3]{E3} for chunking, while others like \hyperref[e:5]{E5} primarily occur as a secondary error along with another type.
Notably, \emph{Fabricated Content} \hyperref[e:10]{(E10)}—often referred to as hallucination—is relatively rare in our analysis. This observation stands in contrast to the focus on hallucination in recent research and public discourse surrounding LLMs \cite{ji2023survey}. Our findings suggest that, in the context of RAG, other types of errors such as retrieval or chunking issues are more prevalent and may warrant greater attention.

\begin{table}[t]
\centering
\small
\caption{Error-type distributions for DragonBall-EN. Mode and 2\textsuperscript{nd} Mode show the original RAG pipeline, while Impr. shows the mode after targeted improvements to the RAG pipeline.}
\label{tab:mode_error_counts}
\setlength{\tabcolsep}{2pt}
\resizebox{\columnwidth}{!}{
\begin{tabular}{lrr|r}
\toprule
\textbf{Error Classification} & \textbf{Mode} & \textbf{2\textsuperscript{nd} Mode} & \textbf{Impr.} \\
\hline
\hyperref[e:1]{E1 Overchunking}             & 0         &    19   & 0        \\
\hyperref[e:2]{E2 Underchunking}            & 32         &     92   & 10     \\
\hyperref[e:3]{E3 Context Mismatch}        & 207       &      32     & 78    \\
\textbf{Percentage Chunking}  &   \textbf{29.7\%}  &     \textbf{32.2\%}    & \textbf{16.4\%}   \\
\hline
\hyperref[e:4]{E4 Missed Retrieval}         &  229       &      27     & 122    \\
\hyperref[e:5]{E5 Low Relevance}            &  33       &       107    & 8    \\
\hyperref[e:6]{E6 Semantic Drift}           &  0     &           7  & 0  \\
\textbf{Percentage Retrieval} &     \textbf{31.5\%}   &  \textbf{31.8\%}      & \textbf{24.3\%}   \\
\hline
\hyperref[e:7]{E7 Low Recall} &  53      &   5 & 28            \\
\hyperref[e:8]{E8 Low Precision} &  6     &    21      & 0     \\
\textbf{Percentage Reranking} &   \textbf{7.1\%}      &    \textbf{5.9\%}    & \textbf{5.2\%}     \\
\hline
\hyperref[e:9]{E9 Abstention Failure}       &  6   &       0    & 5    \\
\hyperref[e:10]{E10 Fabricated Content}       &  34   &       37   & 53     \\
\hyperref[e:11]{E11 Parametric Overreliance}  &  4  &        6     & 0 \\
\hyperref[e:12]{E12 Incomplete Answer}        &  17  &        22    & 26   \\
\hyperref[e:13]{E13 Misinterpretation}        &  178  &         42      & 164           \\
\hyperref[e:14]{E14 Contextual Misalignment}  &  14  &         18        & 8         \\
\hyperref[e:15]{E15 \footnotesize{Chronological Inconsistency}} &  3 &          3   & 13              \\
\hyperref[e:16]{E16 Numerical Error}          &  16   &         6       & 19         \\
\textbf{Percentage Generation}&       \textbf{32.7\%}        &      \textbf{30.2\%}     & \textbf{53.9\%}    \\
\hline
\textbf{Total} & \textbf{832} &  \textbf{444} & \textbf{534}\\
\bottomrule
\end{tabular}
}
\end{table}

Next we demonstrate how insights can be gathered into a RAG pipeline's behaviour through fine-grained analysis with RAGEC. \Cref{tab:error_type_domain_qtypes} shows the distribution of the error types across knowledge domains and query types defined in DragonBall-EN. 

\textbf{Error Types across Knowledge Domains.} We see that \textit{finance} queries lead to the most diversity in error types, but also skew more towards generation errors. Misinterpretation \hyperref[e:13]{(E13)} is the single biggest challenge. For \textit{law} queries three of the RAG stages contribute roughly the same amount of errors, and overall law is the most difficult domain. Numerical errors \hyperref[e:16]{(E16)} are more common for law than any other domain which may reveal an underlying deficiency in the generation model around calculations that are typically done in law, as opposed to finance. \textit{Medical} queries are by far dominated by chunking errors, with context mismatch \hyperref[e:3]{(E3)} having an outsized presence. A practitioner seeing this could simply increase chunk size, return entire documents corresponding to the top chunks, or invest in more complex solutions like expanding chunked text to include explicit definitions.

\textbf{Error Types across Query Types.} 
For \textit{factual questions} three error types dominate (\hyperref[e:3]{E3}, \hyperref[e:4]{E4}, \hyperref[e:13]{E13}), whereas \textit{unanswerable questions}, where the correct response is to abstain, show the highest diversity in error types. As expected, the abstention failure \hyperref[e:9]{(E9)} error type is used nearly exclusively for unanswerable questions.
\textit{Summarization} queries, the most difficult type, are dominated by chunking errors (\hyperref[e:2]{E2}, \hyperref[e:3]{E3}). Multiple semantically related passages are typically needed to produce a coherent and complete summary, and these connections can be lost because of chunking.
\textit{Numerical comparison} queries show fewer generation errors than other types, and numerical errors \hyperref[e:16]{(E16)} are not particularly troublesome. Errors are instead coming from context mismatch \hyperref[e:3]{(E3)} and missed retrieval \hyperref[e:4]{(E4)}.
Finally, as expected, \textit{multi-hop reasoning} is one of the more challenging types and displays a wide range of error types for this RAG implementation. This range accurately reflects the more complicated nature of multi-hop reasoning queries where many things can go wrong, and demonstrates that the proposed error categories are expressive enough to capture the diversity of errors in complex query types.

\begin{table*}[t]
\centering
\small
\caption{Modal error type distributions for DragonBall-EN across knowledge domains and query types. Query type labels are provided by the dataset.}
\label{tab:error_type_domain_qtypes}
\setlength{\tabcolsep}{2pt}
\resizebox{2\columnwidth+18pt}{!}{
\begin{tabular}{lccc|ccccccc}
\toprule
\multirow{2}{*}{\textbf{Error Classification}}&\multicolumn{3}{c|}{\textbf{Knowledge Domain}}& \multicolumn{7}{c}{\textbf{Query Type}}\\
& \textbf{Finance} & \textbf{Law} & \textbf{Medicine} & 
\textbf{\begin{tabular}{@{}c@{}}Factual \\ Question\end{tabular}}
& \textbf{\begin{tabular}{@{}c@{}}Unanswerable \\ Question\end{tabular}} 
& \textbf{\begin{tabular}{@{}c@{}}Numerical \\ Comparison\end{tabular}} & 
\textbf{\begin{tabular}{@{}c@{}}Information \\ Integration\end{tabular}} & 
\textbf{\begin{tabular}{@{}c@{}}Temporal \\ Sequence\end{tabular}} & 
\textbf{\begin{tabular}{@{}c@{}}Multi-hop \\ Reasoning\end{tabular}} & 
\textbf{\begin{tabular}{@{}c@{}}Summari- \\ zation\end{tabular}}\\
\hline
\hyperref[e:1]{E1 Overchunking}             & 0         &    0   & 0  & 0 & 0 & 0 & 0 & 0 & 0 & 0      \\
\hyperref[e:2]{E2 Underchunking}            & 2         &     19   & 11 & 1 &3&	1&0&1&7&19    \\
\hyperref[e:3]{E3 Context Mismatch}        & 19       &      71     & 117 &19 &5 &22 &43 &12 &24 &82\\
\textbf{Percentage Chunking}  &\textbf{5.9\%}&\textbf{32.8\%}&\textbf{62.4\%}&\textbf{36.4\%}&\textbf{16.3\%}&\textbf{29.1\%}&\textbf{18\%}&\textbf{14.6\%}&\textbf{27\%}&\textbf{49\%}  \\
\hline
\hyperref[e:4]{E4 Missed Retrieval}         &  94       &      76     & 59 & 22 &7 &29 &87 &35 &20 &29 \\
\hyperref[e:5]{E5 Low Relevance}            &  25       &       6    & 2 & 1 &0 &4 &20 &1 &3 &4 \\
\hyperref[e:6]{E6 Semantic Drift}           &  0     &           0  & 0 &0&0&0&0&0&0&0 \\
\textbf{Percentage Retrieval} &\textbf{33.7\%}&\textbf{29.9\%}&\textbf{29.8\%}&\textbf{41.8\%}&\textbf{14.3\%}&\textbf{41.8\%}&\textbf{44.8\%}&\textbf{40.4\%}&\textbf{20\%}&\textbf{16\%}  \\
\hline
\hyperref[e:7]{E7 Low Recall} &  28      &   23 & 2      & 0 &2 &3 &25 &10 &8 &5 \\
\hyperref[e:8]{E8 Low Precision} &  4     &   2      & 0   &1&1&2&0&1&0&1  \\
\textbf{Percentage Reranking} &\textbf{9.1\%}&\textbf{9.1\%}&\textbf{1\%}&\textbf{1.8\%}&\textbf{6.1\%}&\textbf{6.3\%}&\textbf{10.5\%}&\textbf{12.4\%}&\textbf{7\%}&\textbf{2.9\%}    \\
\hline
\hyperref[e:9]{E9 Abstention Failure}       &  5   &       1    & 0  &0&5&0&1&0&0&0  \\
\hyperref[e:10]{E10 Fabricated Content}       &  30   &       4   & 0  &0&4&2&5&0&8&15   \\
\hyperref[e:11]{E11 Parametric Overreliance}  &  4  &        0     & 0 &0&0&0&0&0&1&3\\
\hyperref[e:12]{E12 Incomplete Answer}        &  8  &        8    & 1 &0&0&0&7&0&8&2  \\
\hyperref[e:13]{E13 Misinterpretation}        &  116  &         50      & 12  &10&16&11&46&29&26&40         \\
\hyperref[e:14]{E14 Contextual Misalignment}  &  11  &         2        & 1  &0&4&2&3&0&0&5       \\
\hyperref[e:15]{E15 \footnotesize{Chronological Inconsistency}} &  2 &          1   & 0 &1&0&0&1&0&0&1             \\
\hyperref[e:16]{E16 Numerical Error}          &  5   &        11      & 0  &0&2&3&1&0&10&0       \\
\textbf{Percentage Generation}&\textbf{51.3\%}&\textbf{28.1\%}&\textbf{6.8\%}&\textbf{20\%}&\textbf{63.3\%}&\textbf{22.8\%}&\textbf{26.8\%}&\textbf{32.6\%}&\textbf{46.1\%}&\textbf{32\%}   \\
\hline
\textbf{Total Errors} &\textbf{353}&\textbf{274}&\textbf{205}&\textbf{55}&\textbf{49}&\textbf{79}&\textbf{239}&\textbf{89}&\textbf{115}&\textbf{206}\\
\textbf{Total Queries} &\textbf{1574}&\textbf{782}&\textbf{752}&\textbf{535}&\textbf{233}&\textbf{297}&\textbf{737}&\textbf{241}&\textbf{532}&\textbf{533}\\
\textbf{Error Rates} &\textbf{22.4\%}&\textbf{35\%}&\textbf{27.3\%}&\textbf{10.3\%}&\textbf{21\%}&\textbf{26.6\%}&\textbf{32.4\%}&\textbf{36.9\%}&\textbf{21.6\%}&\textbf{38.6\%}\\
\bottomrule
\end{tabular}
}
\end{table*}

\begin{table}[t]
\centering
\small
\caption{Distribution of mode frequency ($K=10$).}
\label{tab:mode_freq_categories}
\setlength{\tabcolsep}{3pt}
\begin{tabular}{ccc}
\toprule
\textbf{Mode Frequency} & \textbf{DragonBall} & \textbf{CLAPnq}\\
\midrule
3&1&4\\
4&7&32\\
5&41&91\\
6&74&123\\
7&74&135\\
8&94&149\\
9&153&172\\
10&388&516\\
\bottomrule
\end{tabular}
\end{table}

\textbf{Classification Consistency.} Finally, in \Cref{tab:mode_freq_categories} we examine the consistency of classifications via mode frequency, the number of the $K$ runs that agreed with the modal error type. For the majority of the queries, RAGEC is quite consistent in determining error categories. \Cref{app:co-occurences} discusses the co-occurrence of the error categories.

\textbf{Summary.} RAGEC provides an entry point for debugging and improvement of RAG pipelines, which can be complex and opaque to developers. Instead of manually sifting through individual generations, our system automatically captures intermediate information like retrieved and reranked chunks, and distills the information into high-level descriptive statistics. Hence, our focus is not on numerical metrics of performance, but on understanding the types and causes of errors so that developers can intelligently prioritize aspects to improve.

\textbf{Using RAGEC to Improve RAG.}
We conduct a case study on how RAGEC can be used in practice to improve a RAG pipeline. We focus on DragonBall-EN where 832 errors were originally identified (73.3\% correct). RAGEC indicates that chunking, retrieval, and generation all caused roughly the same proportion of errors (Table~\ref{tab:mode_error_counts}, Mode), and hence are the primary candidates for improvement. Since most chunking errors come from Context Mismatch \hyperref[e:3]{(E3)} where contextual links are broken across chunks, we follow the recommendation in \Cref{sec:chunking} and modify the chunking strategy from fixed-length segmentation to a recursive, sentence-level segmentation of approximately the same size. In addition, RAGEC indicates that most retrieval and reranking errors come from Missed Retrieval \hyperref[e:4]{(E4)} and Low Recall \hyperref[e:7]{(E7)} which are addressed by increasing the number of chunks retrieved and passed to the generator. 

With these improvements, we reran the RAG pipeline and RAGEC. The improved pipeline resulted in fewer errors overall, only 534 (82.8\% correct). In Table~\ref{tab:mode_error_counts} (column Impr.) we observe that chunking-, retrieval-, and reranker-related errors were signficantly reduced, while generator errors stay similar as no intervention was done.

\section{Conclusion}\label{sec:conclusion}

Our contributions in this work provide a framework for practitioners working with RAG systems to understand, categorize, track, and fix model errors. Our findings also indicate the wide range of possible avenues for improving the robustness of RAG systems in future works. While our RAG error classification method, RAGEC, achieves a useful level of agreement to human error annotations, it is far from perfect, indicating that error classification remains a challenging problem for LLM-based systems. To promote future research in this area, we release our annotated dataset of error types as a resource for the community.

\section*{Limitations}

Our RAG error taxonomy is more expansive than prior works in this vein, but still is not exhaustive of all possible errors that could occur, especially if further stages were to be added to the RAG pipeline. We focused on the most prevalent types observed in practice.

The RAG pipeline we used for demonstrating auto-evaluation for error classification reflects the general architecture of modern pipelines used in real-world applications. However, it was not highly tuned for the DragonBall and CLAPnq datasets, for example by implementing the many improvements we listed in \Cref{sec:classification}. Our aim was to show a starting point, where practitioners can understand errors in their system before making changes.

Additionally, our study focuses on single-turn textual queries. Expanding the evaluation framework to handle multi-turn conversations or multimodal inputs remains an important direction for future work.

Finally, we found RAG error classification according to our taxonomy to be a challenging prediction problem. Part of this stems from the difficulty of collecting labeled data. The error type labels we collected require annotators to be experts in RAG systems so that they comprehend the role and operation of each stage, and can disambiguate the error types. Hence, crowdsourcing the annotation work was not possible, and all annotation was done manually by the authors to ensure the highest quality labels. Still, LLM-as-a-Judge systems were not fully capable of reasoning over all the intermediate information created throughout the RAG pipeline, leading to rather low stage-classification accuracy. We expect this to increase as LLMs become more proficient with multi-step reasoning and assimilating information over long contexts.

\bibliography{custom}

\clearpage
\appendix

\section{RAG Pipeline and Dataset Details}
\label{app:a}

\vspace{-2pt}
This section outlines the configurable components of our RAG pipeline and provides details on the specific configurations used to obtain the experimental results presented in the main paper. In \Cref{tab:rag_hyperparam}, we list important hyperparameters used on the DragonBall and CLAPnq datasets. Note that the DragonBall dataset is publicly available under a CC BY-NC-SA 4.0 license. The CLAPnq dataset is publicly available under an Apache 2.0 license.

\vspace{-2pt}
\paragraph{Chunker.}
We implemented multiple chunking strategies, including fixed-length chunking (with and without overlap) and recursive chunking at the sentence level. These methods enable flexible preprocessing of documents depending on the context granularity needed. When available, our implementation also handles pre-defined semantic chunks provided by the dataset. For instance, the CLAPnq dataset provides pre-chunked documents aligned with natural discourse boundaries, which we use directly in our experiments. For the DragonBall dataset, we follow the original paper’s best configuration and apply fixed-length chunking with a window size of 128 tokens and an overlap of 25 tokens \citep{zhu2024rageval}.

\vspace{-2pt}
\paragraph{Retrieval.}  
For dense retrieval, we used models from the Hugging Face repository (e.g., \texttt{gte-large-en-v1.5}) to generate embeddings of the queries and chunks. Retrieval computes vector similarity (using cosine similarity) between the query and document chunks and selects the top-$k$ most relevant candidates. The value of $k$ is treated as a hyperparameter. We fixed $k=8$ and $k=5$ for the DragonBall and CLAPnq datasets respectively.

\vspace{-2pt}
\paragraph{Reranker.}  
For the reranking stage, we use specialized LLM reranker models fine-tuned for ranking tasks (e.g., \texttt{rank-zephyr-7b-v1-full}), also sourced from Hugging Face. Given a set of retrieved candidate chunks and the original query, this model assigns a relevance score to each chunk. We then select the top $k'$ candidates (with $k' < k$ from retrieval) to ensure that only the most relevant context is passed to the generator. The reranking step plays a critical role in filtering noisy or marginally relevant content returned by dense retrieval. This two-stage process is meant to enhance retrieval precision, thereby improving final answer quality. We fixed $k'=5$ and $k'=3$ for the DragonBall and CLAPnq datasets respectively. The system prompt is reproduced in Listing~\ref{lst:system-prompt-rag-reranker}.

\begin{lstlisting}[caption=System Prompt Used for our RAG Reranker,label=lst:system-prompt-rag-reranker]
"""
System:
You are RankLLM, an intelligent assistant that can rank passages based on their relevancy to the query.

User:
I will provide you with {{K}} passages, each indicated by a numerical identifier []. Rank the passages based on their relevance to the search query: {{query}}

[1] {{chunk1}}
[2] {{chunk2}}
...
[K] {{chunkK}}

Search Query: {{query}}. 

Rank the {{K}} passages above based on their relevance to the search query. All the passages should be included and listed using identifiers, in descending order of relevance. The output format should be [] > [], e.g., [2] > [1]. Only respond with the ranking results, do not say any word or explain.

Assistant:
"""
\end{lstlisting}

\paragraph{Generator.}  
The final stage involves generating answers using an LLM. The reranked top $k'$ chunks are concatenated and provided as context to the model alongside the query. Note that only the retrieved chunks (not full documents) are used as context. Our implementation is modular enough so that one can experiment with several LLMs from Hugging Face and select the one that best suits the needs. In our experiments, we used \texttt{Meta-Llama-3-8B-Instruct}. The system prompt is reproduced in Listing~\ref{lst:system-prompt-rag-generator}.

\begin{table}[t]
\centering
\small
\caption{RAG Implementation Hyperparameters}
\label{tab:rag_hyperparam}
\vspace{-4pt}
        \setlength{\tabcolsep}{2pt} 
\begin{tabular}{lll}
\toprule
\textbf{Parameter} & \textbf{DragonBall} & \textbf{CLAPnq} \\
\midrule
Chunker    &    Fixed-Length  &       Semantic      \\
& size=128, overlap=25 & \\
Embedder   &    gte-large-en-v1.5      &     gte-large-en-v1.5        \\
Search     &    Top-$k$=8      &        Top-$k$=5     \\
Reranker   &    rank-zephyr-7b-v1-full  & rank-zephyr-7b-v1-full \\
 & Top-$k'$=5 & Top-$k'$=3 \\
Generator  &    Llama-3-8B-Instruct      &        Llama-3-8B-Instruct      \\
\bottomrule
\end{tabular}
\vspace{-16pt}
\end{table}

\begin{lstlisting}[caption=System Prompt Used for our RAG Generator,label=lst:system-prompt-rag-generator]
"""
System:
You are an assistant for answering queries. You are given a list of context (extracted parts of some documents) and a query. Based on the given context, provide an answer to the query. Please be concise and to the point. If you don't know the answer say 'I don't know!' Don't make up an answer. Cite the document id used. The output format should be answer with citations. Only respond with the answer, do not explain.

User:
Query: 
Where is the capital of France?
Context: 
{document id: 1, content: Paris is the capital of France, the largest country of Europe with 550 000 km2).}{document id: 2, content: France is a country in Europe.}

Assistant:
Answer: Paris. [1]

User:
Query:
{{query}}
Context:
{{Context}}
"""
\end{lstlisting}

\section{Automatic Error Classification Details}
\label{app:b}

This section provides additional materials that support our main auto-evaluation analysis from \cref{sec:auto-eval}. We begin with an extended description of our auto-evaluation pipeline, RAGEC, including details such as the system prompts we used. Additionally, we present alternative auto-evaluation methods which empirically had lower agreement with our human annotations.

\subsection{Answer Evaluation System Prompts}\label{app:auto-eval-prompts}

The first step of RAGEC, answer evaluation uses an LLM to determine if a generated response is correct, given the query, and ground-truth answer and documents. In practice, we used GPT-4o with the prompt in Listing~\ref{lst:system-prompt-gpt4o}.

\begin{lstlisting}[caption=System Prompt Used for Answer Evaluation,label=lst:system-prompt-gpt4o]
    """
    You are an expert evaluator. Your task is to evaluate if a proposed answer matches the ground truth answer for a given question.

    You will be provided with information between special tags:
    1. The original question <question>
    2. The ground truth (correct) answer <ground_truth>
    3. A proposed answer to evaluate <proposed_answer>
    4. (Optional) the proposed answer's cited information <ground_truth_citations>

    Please evaluate the proposed answer based on the following criteria:
    - Abstain: If the proposed answer is "I don't know" or "I don't have enough information to answer this question", then the label is 'abstain'.
    - Accuracy: Does it contain the same key information as the ground truth?
    - Completeness: Does it cover all important points from the ground truth?
    - Correctness: Are there any factual errors compared to the ground truth?

    Provide your evaluation as a JSON object with the following fields:
    {{
        \"label\": \"correct\" | \"possible_correct\" | \"incorrect\" | \"abstain\",
        \"reasoning\": string // A very brief explanation of your evaluation
    }}

    Here are some examples:

    <...>

    <question>
    {question}
    </question>

    <ground_truth>
    {ground_truth}
    </ground_truth>

    <proposed_answer>
    {proposed_answer}
    </proposed_answer>

    {citations}

    Your evaluation:
    """
\end{lstlisting}

\subsection{Error Stage Classification}
This section describes the algorithm for the identification of which stage in the RAG pipeline caused the error.

\subsubsection{Determining the ground truth chunks}
The first step is to determine which chunks contain the ground truth information from the corpus. For CLAPnq, the dataset provides short ground truth chunks along with each query; thus we use the dataset's chunks directly.

For DragonBall, only the document containing the ground truth is given with the query, but these documents are long, so we perform the chunking ourselves. We leverage an LLM to determine which chunks are necessary to answer the query. We provide GPT-4o-mini the query, ground truth answer, and all the chunks from the ground truth documents, and ask for the IDs of chunks that contain information necessary to answer the query. We perform this process 10 times and select chunks appearing more than 8 times as ground truth chunks. The repetition accounts for the stochasticity of using an LLM, and better reflects its confidence. The prompt is shown below in Listing~\ref{lst:system-prompt-gt-chunk}.

\begin{lstlisting}[caption=System prompt for determining the chunks containing the ground truth on the DragonBall dataset,label=lst:system-prompt-gt-chunk]
"""
You are an expert evaluator for Retrieval Augmented Generation (RAG) systems. You will help identify chunking-related errors in RAG systems by analyzing the relationship between queries, ground truth answers, and document chunks. Your task is to identify which chunks from the ground truth document are relevant to answering the query.

## Instructions
1. Review ALL available chunks from the ground truth document
2. Identify ALL chunks containing information relevant to answering the query
3. Consider both direct and indirect relevance to the query
4. Select chunks that together provide sufficient information to answer the query

## Context
**Query**: {query}
**Ground Truth Answer**: {ground_truth}

**Available Chunks from Ground Truth Document**:
{chunks}

## Output Format
Provide your answer in the following format:
Relevant Chunks: [45_1, 45_4, 45_10]

"""
\end{lstlisting}

\subsubsection{Determining errors in the generation stage}
Given the ground truth chunks, if we know that the chunks being passed to the generator are sufficient to determine the answer, we can conclude the error happened in the generation stage. Thus we look at the fraction of the ground truth chunks being passed to the generator. However, we must account for the possibility that multiple chunks contain the ground truth. In these cases, using only a subset of the ground truth chunks may be sufficient to determine the correct answer. To this end, we specify that if more than half of the ground truth chunks are passed to the generator, it is deemed a generator error.

If there are no ground truth chunks extracted, it is likely that the error is an abstention error as the question may be unanswerable. Thus it is also deemed a generator error in this case.

\subsubsection{Determining errors in the reranking stage}
An error occurs in the reranking stage if the reranker filters out some ground truth chunks which the retriever had returned. Thus if there is any ground truth chunk that is filtered out at this stage, we deem the error a reranker error. 

Note that it could be the case that the reranker filters out redundant chunks containing the ground truth, as described in the above step. However, in practice, if both chunks containing the correct answer to the question are retrieved, it is not likely that the reranker would treat them differently. We found the described method to be the most effective in determining whether there is a reranker error.

\subsubsection{Distinguishing between chunking errors and retrieval errors}
Chunking errors and retrieval errors are difficult to distinguish, as the efficacy of the retriever is tightly coupled to the quality of chunks. Chunking error types \hyperref[e:1]{E1}, \hyperref[e:2]{E2}, and \hyperref[e:3]{E3} all relate to cases where important information is missing or overshadowed in the chunks. Thus we aim to check if each ``concept'' in each query is preserved in at least one of the chunks containing the ground truth.

To extract the concepts in each query, we follow closely the method from \citet{walk-the-talk}. We ask GPT-4o-mini to extract the concepts from each query, then for each concept, we ask the LLM which ground truth chunks contain that concept. We then compute the fraction of all concepts which are included in at least one of the ground truth chunks. If this fraction is less than 0.8, we deem the error to be a chunking error. Otherwise it is deemed a retrieval error.

The parameter 0.8 was chosen to reflect that if the number of concepts extracted from the query is high, it is likely that not all the concepts are needed to answer the question. Thus we allow some leeway for the cases where the number of concepts is high (namely greater than 6).

The prompt for determining the concepts in the query is given in Listing~\ref{lst:system-prompt-concept}, and for determining whether a concept is contained in the ground truth chunks is given in Listing~\ref{lst:system-prompt-concept-validator}.

\begin{lstlisting}[caption=System prompt used for determining the concepts in each query,label=lst:system-prompt-concept]
"""
Consider the following questions. Your task is to list the set of distinct concepts, or high-level pieces of information, in the 'Context' that could possibly influence someone's answer to the question. Each concept should appear word-to-word in the question, or a very minor rewording. Here are three examples.
Example 1:
Question:
Compare the debt restructuring efforts of Company A in 2018 and Company B in 2021. Which company reduced more liabilities through debt restructuring?

Concept List:
Debt restructuring efforts
Company A
2018
Company B
2021
Reducing liabilities through debt restructuring

{examples 2 and examples 3}

Please fill out the 'Concept List' for the fourth example by providing a numbered list. You should not restate the 'Concept List' header. You should not put dash ('-') or numbers before each item in the list.
Example 4
{query}
"""
\end{lstlisting}

\begin{lstlisting}[caption=System prompt used for determining whether a concept is contained in the ground truth chunks,label=lst:system-prompt-concept-validator]
"""
You will be given a list of excerpts and a concept. Your job is to determine whether the concept given is contained in each excerpt. Output a line for excerpt, output "True" or "False".

Example 1:
**Concepts**: Peter
**Excerpts**:
[45_3] John is running
[45_6] Peter is walking
**Answer**:
[45_3] False
[45_6] True

Example 2:
**Concepts**: running
**Excerpts**:
[45_3] John is running
[45_6] Peter is walking
**Answer**:
[45_3] True
[45_6] False

Question:
**Concepts**: {concept}
**Excerpts**:
{excerpts}
**Answer**:
"""
\end{lstlisting}

\subsection{Error Type Classification}
After determining the RAG stage responsible for the error, we then prompt GPT-4o-mini $K=10$ times to judge the error type, and use the distribution of results to evaluate the LLM's confidence. Each stage uses a different system prompt and different contextual information relevant to that stage.

\subsubsection{Chunking error type}
If stage classification determined the error to be a chunking error, then classifying which type of chunking error only requires the query, ground truth answer, and chunks from the ground truth document. The prompt for chunking errors is given in Listing~\ref{lst:system-prompt-chunking}. 

\begin{lstlisting}[caption=System prompt used for determining chunking error types,label=lst:system-prompt-chunking]
"""
You are an expert Retrieval Augmented Generation (RAG) system evaluator.
Background:
Given a query, and a list of documents possibly containing the answer to the query, a RAG model has tried to answer the query using the following 4 steps.
- Chunking: Each document in the document list is split into smaller chunks.
- Retrieval: A specified number of chunks closest to the query will be retrieved.
- Reranking: The retrieved chunks are ranked for the second time, according to how relevant the chunks are to the query. Then, only a specified number of reranked chunks are retained.
- Generation: The query and reranked chunks are passed to the generator LLM to generate an answer.

You will be given a query and the ground truth answer, as well as all the chunks that belong to the document containing the ground truth answer. As described in the background, each chunk will be seen as independent text blocks by the RAG model. Given there is an error in the chunking step, your job is to determine the best description of the error from the list below.
- Overchunking: Document is split into excessively small chunks, causing important context to be lost. Individual chunks appear incomplete or ambiguous.
- Underchunking: Chunks are too large, covering multiple topics with mixed content. Individual chunks can be confusing and crucial information is diluted.
- Context Mismatch: Chunks are split at arbitrary boundaries, disrupting the logical structure of the document. Key contextual links are separated from the information they link to

Context:
**Query**: {query}
**Ground Truth**: {ground_truth}
**Document Chunks**:
{ground_truth_chunks}

**Output format**:
Your answer should only contain one of the following,
Overchunking, Underchunking, Context Mismatch
"""
\end{lstlisting}

\subsubsection{Retrieval error type}
For retrieval errors, the LLM only needs the query, the ground truth answer, the ID for the ground truth documents, and the retrieved chunks. The prompt for retrieval errors is given in Listing~\ref{lst:system-prompt-retrieval}. 

\begin{lstlisting}[caption=System prompt used for determining retrieval error types,label=lst:system-prompt-retrieval]
"""
You are an expert Retrieval Augmented Generation (RAG) system evaluator.
Background:
Given a query, and a list of documents possibly containing the answer to the query, a RAG model has tried to answer the query using the following 4 steps.
- Chunking: Each document in the document list is split into smaller chunks.
- Retrieval: A specified number of chunks closest to the query will be retrieved.
- Reranking: The retrieved chunks are ranked for the second time, according to how relevant the chunks are to the query. Then, only a specified number of reranked chunks are retained.
- Generation: The query and reranked chunks are passed to the generator LLM to generate an answer.

You will be given a query and the ground truth answer. You will also be given IDs of the documents containing the ground truth and a selection of document chunks retrieved.
Given there is an error in the retrieval step, your job is to determine the best description of the error from the list below.
- Missed Retrieval: Retrieved chunks do not contain the relevant information to answer the query from the ground truth documents
- Low Relevance: Retrieved chunks are only loosely related to the query
- Semantic Drift: Retrieved chunks appear to match keywords but do not align with the query's intent

Context:
**Query**: {query}
**Ground Truth**: {ground_truth}
**Ground Truth Document ID**: {ground_truth_doc_ids}
**Retrieved Chunks**:
{retrieved_chunks}

**Output format**:
Your answer should only contain one of the following:
Missed Retrieval, Low Relevance, Semantic Drift
"""
\end{lstlisting}

\subsubsection{Reranking error type}
To determine the reranking error type, the LLM only needs the query, the ground truth answer, the retrieved chunks, and the reranked chunks. The prompt for reranking error is given in Listing~\ref{lst:system-prompt-rerank}. 

\begin{lstlisting}[caption=System prompt used for determining reranking error types,label=lst:system-prompt-rerank]
"""
You are an expert Retrieval Augmented Generation (RAG) system evaluator.
Background:
Given a query, and a list of documents possibly containing the answer to the query, a RAG model has tried to answer the query using the following 4 steps.
- Chunking: Each document in the document list is split into smaller chunks.
- Retrieval: A specified number of chunks closest to the query will be retrieved.
- Reranking: The retrieved chunks are ranked for the second time, according to how relevant the chunks are to the query. Then, only a specified number of reranked chunks are retained.
- Generation: The query and reranked chunks are passed to the generator LLM to generate an answer.

You will be given a query and the ground truth answer. You will also be given a selection of document chunks retrieved. The retrieved chunks will be reranked so that only {num_reranked_chunks} chunks are further selected. Given there is an error in the reranking step, your job is to determine the best description of the error from the list below.
- Low Recall: Necessary chunks are retrieved but reranked too low and not forwarded to the generator
- Low Precision: Irrelevant chunks are reranked highly and forwarded to the generator, with greater importance than the truly relevant chunks

Context:
**Query**: {query}
**Ground Truth**: {ground_truth}
**Ground Truth Document ID**: {ground_truth_doc_ids}
**Reranked Chunks**:
{reranked_chunks}

**Output format**:
Your answer should only contain one of the following:
Low Recall, Low Precision
"""
\end{lstlisting}

\subsubsection{Generation error type}
To determine the generation error type, the LLM only needs the query, the ground truth answer, incorrect answer by RAG model and the reranked chunks. The prompt for generation error types is as follows. 

\begin{lstlisting}[caption=System prompt psed for determining generation error types,label=lst:system-prompt-generation]
"""
You are an expert Retrieval Augmented Generation (RAG) system evaluator.
Background:
Given a query, and a list of documents possibly containing the answer to the query, a RAG model has tried to answer the query using the following 4 steps.
- Chunking: Each document in the document list is split into smaller chunks.
- Retrieval: A specified number of chunks closest to the query will be retrieved.
- Reranking: The retrieved chunks are ranked for the second time, according to how relevant the chunks are to the query. Then, only a specified number of reranked chunks are retained.
- Generation: The query and reranked chunks are passed to the generator LLM to generate an answer.

You will be given a query, the ground truth answer and an incorrect answer to the query generated by the RAG model. You will also be given {num_reranked_chunks} document chunks. Your job is to determine the reason why the model outputs the incorrect answer, from the list below.
- Abstention Failure: The model should have abstained but provided an incorrect answer
- Fabricated Content: The response includes information not present in the source document chunks and is unverifiable
- Parametric Overreliance: The response depends on the LLM's internal knowledge rather than the source document chunks
- Incomplete Answer: The response is correct but missing critical details
- Misinterpretation: The generator misuses or misrepresents the source document chunks
- Contextual Misalignment: The response is correct but does not address the query
- Chronological Inconsistency: The response presents events or facts in the wrong temporal order, or confuses past, present, or future timelines
- Numerical Error: The response includes incorrect calculations, quantities, or misrepresents numerical data from the retrieved documents

Context:
**Query**: {query}
**Ground Truth**: {ground_truth}
**Incorrect Answer**: {incorrect_rag_answer}
**Reranked Chunks**:
{reranked_chunks}

**Output format**:
Your answer should only contain one of the following:
Abstention Failure, Fabricated Content, Parametric Overreliance, Incomplete Answer, Misinterpretation, Contextual Misalignment, Chronological Inconsistency, Numerical Error
"""
\end{lstlisting}

\subsection{Alternative Auto-evaluation Approaches}\label{app:alternatives}

So far this appendix has described in detail RAGEC, our proposed error classification method. While designing RAGEC, we explored many other systems for LLM-based error classification, but RAGEC empirically had the best agreement rates with our human-annotated data. For comparison, in this section we describe two alternative approaches to error classification that we tested. For each alternative, we used GPT-4o-mini, the same model as RAGEC above for fair comparison.

\subsubsection{Single-step error type classification}

The most straightforward approach to error type classification is to directly prompt an LLM to output one of the 16 defined error types given all available contextual information, including the query, ground truth response from the dataset, generated response, retrieved chunks, and reranked chunks. Consistently with other methods, the 16 error types are defined and described in the system prompt. We apply this method directly after the \emph{answer evaluation} step of RAGEC, such that the same set of 832 potential errors from DragonBall are annotated.

Compared to RAGEC which achieved 57.8\% stage classification agreement, and 40.3\% error-type classification agreement, single-step prompting had only 41.1\% and 31.1\% agreement rates, respectively. Note that stage classification is done implicitly, taking the stage of the modal error type. Single-step prompting overloads the LLM with information which must be digested and synthesized in a single shot. Breaking the classification up by stage helps to guide the LLM's reasoning towards a smaller set of possibilities, potentially with a more focused collection of contextual information.

\subsubsection{Stage-sequential error type classification}

As above, we begin with the \emph{answer evaluation} stage of RAGEC to identify which RAG responses need to be classified. After this, we proceed sequentially over the stages of the RAG pipeline in order, starting with chunking. At each stage, an LLM is prompted with context to determine if an error occurred at this stage, and if so, which of the error types occurred within that stage. Once the earliest error is identified, the evaluation stops.

In more detail, for the chunking stage, the context contains the query, ground truth response from the dataset, and all chunks from the ground truth document. Notably, we do not include information like the generated response or retrieved chunks, because these are not available during the chunking step and hence are not relevant to determining chunking errors. Next, for the retrieval stage we provide the query and ground truth response, but now include the retrieved chunks and the ground truth document ID. Continuing to reranking if necessary, the retrieved chunks are replaced by the reranked chunks. Because of the sequential nature of the pipeline, retrieved chunks are not strictly necessary here. Finally, if no error is identified in the first three stages, we assume the generator caused the error by process of elimination. The LLM is prompted to classify the error type given the query, ground truth response, generated response, and reranked chunks.

Stage-sequential error type classification performed better than single-step with 47.5\% stage classification agreement, and 36.3\% error-type classification, but this still falls short of RAGEC by a considerable margin. We analysed the performance of stage-sequential error type classification compared to human annotations and found that it lacked the ability to correctly classify any chunking errors. Without context on the generated response or how chunks were used downstream in the RAG pipeline, determining that chunking was the primary culprit is extremely difficult. This informed how we designed RAGEC, which also iterates over stages sequentially, but in a backward fashion, starting from generation and proceeding back to chunking. We also provided RAGEC more information for the chunking stage by generating ground truth chunk labels with concept extraction.

\section{Additional Results}
\label{app:c}

In this section we present extended results on the co-occurrence patterns of error types, which offer further insight into the structure and dependencies among different failure modes observed in RAG systems beyond what was discussed in \Cref{sec:auto-eval}.

\subsection{CLAPnq Error Type Distribution}\label{app:clapnq-distribution}

Dataset characteristics can influence the failure modes of RAG systems. In Table~\ref{tab:mode_error_counts_clapnq}, we present the distribution of the most and second most frequent error types as identified by RAGEC on the CLAPnq dataset. Compared to the DragonBall-EN dataset (see Table~\ref{tab:mode_error_counts} in \Cref{sec:auto-eval}), the CLAPnq error distribution is much more heavily weighted towards generation errors, with few reranking errors. The lack of chunking errors is because the dataset comes pre-chunked with ground truth properly included.

\begin{table}[t]
\centering
\small
\caption{Distribution of most and second most common error types across the CLAPnq dataset.}
\label{tab:mode_error_counts_clapnq}
\setlength{\tabcolsep}{3pt}
\begin{tabular}{lrr}
\toprule
\textbf{Error Classification} & \textbf{Mode} & \textbf{Second Mode} \\
\hline
\hyperref[e:1]{E1 Overchunking}                      &    0       &   0 \\
\hyperref[e:2]{E2 Underchunking}                     &     0       &  0\\
\hyperref[e:3]{E3 Context Mismatch}                &      0        & 0\\
\textbf{Percentage Chunking}    & \textbf{0\%}         & \textbf{0\%}\\
\hline
\hyperref[e:4]{E4 Missed Retrieval}               &      180        & 69\\
\hyperref[e:5]{E5 Low Relevance}                   &       84   &    99 \\
\hyperref[e:6]{E6 Semantic Drift}                &           1   & 16\\
\textbf{Percentage Retrieval}  &   \textbf{21.69\%}      & \textbf{26.10\%}\\ 
\hline
\hyperref[e:7]{E7 Low Recall} &           23 &          1 \\
\hyperref[e:8]{E8 Low Precision} &       1        &   3\\
\textbf{Percentage Reranking} &    \textbf{1.96\%} & \textbf{0.57\%}    \\
\hline
\hyperref[e:9]{E9 Abstention Failure}       &       9  &   23   \\
\hyperref[e:10]{E10 Fabricated Content}       &      87  &   125  \\
\hyperref[e:11]{E11 Parametric Overreliance}  &       46 &    98 \\
\hyperref[e:12]{E12 Incomplete Answer}        &    51 &    38  \\
\hyperref[e:13]{E13 Misinterpretation}        &    607 &        136        \\
\hyperref[e:14]{E14 Contextual Misalignment}  &         112 &     87           \\
\hyperref[e:15]{E15 Chronological Inconsistency}  &          4 &   6             \\
\hyperref[e:16]{E16 Numerical Error}             &         17  &      4        \\
\textbf{Percentage Generation}               &  \textbf{76.35\%}          & \textbf{73.33\%} \\
\hline
\textbf{Total}  &  \textbf{1222}  & \textbf{705}\\
\bottomrule
\end{tabular}
\end{table}

\begin{table}[t]
\centering
\small
\caption{Distribution of most and second most common error types across the DragonBall-CN dataset.}
\label{tab:mode_error_counts_chinese_dragonball}
\setlength{\tabcolsep}{3pt}
\begin{tabular}{lrr}
\toprule
\textbf{Error Classification} & \textbf{Mode} & \textbf{Second Mode} \\
\hline
\hyperref[e:1]{E1 Overchunking}                      &    1       &   27 \\
\hyperref[e:2]{E2 Underchunking}                     &     8       &  75\\
\hyperref[e:3]{E3 Context Mismatch}                &      185        & 9\\
\textbf{Percentage Chunking}    & \textbf{15.11\%}         & \textbf{18.32\%}\\
\hline
\hyperref[e:4]{E4 Missed Retrieval}               &      296        & 19\\
\hyperref[e:5]{E5 Low Relevance}                   &       25   &    89 \\
\hyperref[e:6]{E6 Semantic Drift}                &           0   & 6\\
\textbf{Percentage Retrieval}  &   \textbf{25.00\%}      & \textbf{18.81\%}\\ 
\hline
\hyperref[e:7]{E7 Low Recall} &           378 &          19 \\
\hyperref[e:8]{E8 Low Precision} &       21        &   155\\
\textbf{Percentage Reranking} &    \textbf{31.07\%} & \textbf{28.71\%}    \\
\hline
\hyperref[e:9]{E9 Abstention Failure}       &       16  &   15   \\
\hyperref[e:10]{E10 Fabricated Content}       &      93  &   57  \\
\hyperref[e:11]{E11 Parametric Overreliance}  &       0 &    7 \\
\hyperref[e:12]{E12 Incomplete Answer}        &    3 &    5  \\
\hyperref[e:13]{E13 Misinterpretation}        &    213 &        77        \\
\hyperref[e:14]{E14 Contextual Misalignment}  &         30 &     42           \\
\hyperref[e:15]{E15 Chronological Inconsistency}  &          4 &   1             \\
\hyperref[e:16]{E16 Numerical Error}             &         11  &      3        \\
\textbf{Percentage Generation}               &  \textbf{28.82\%}          & \textbf{34.16\%} \\
\hline
\textbf{Total}  &  \textbf{1284}  & \textbf{705}\\
\bottomrule
\end{tabular}
\end{table}

\subsection{DragonBall-CN Error Type Distribution}\label{app:chinese-dragonball}
To demonstrate that our method extends to languages other than English, we tested our method on the Chinese subset of the DragonBall dataset. Note that the documents and the queries are in Simplified Chinese. As in \Cref{tab:rag_hyperparam}, the parameters used for chunking, embedding, retrieval, reranking, and generation are all the same as DragonBall-EN, with the exception that chunking uses 512 characters instead of 128 tokens, which is more suitable for the structure of Chinese. We also express the generator prompt in Chinese, as listed below:

\begin{CJK*}{UTF8}{gbsn}
\noindent
\begin{minipage}{\linewidth}
\begin{verbatim}
在此任务中，你需要总结出回答问题所必需的Key Points，并使用这些Key Points来帮助你回答问题。
请按以下格式列出Key Points：
1. ...
2. ...
依此类推，根据需要增加序号,但不超过10个。
每个Key Points必须附带引用来源。随后，利用这些Key Points，在“Answer:”之后生成最终答案。
最终答案中也要引用来源。
\end{verbatim}
\end{minipage}

\noindent
\begin{minipage}{\linewidth}
\begin{verbatim}
确保每个Key Point覆盖不同的内容。
所有答案必须使用中文进行回答。
示例问题：
新修订的《公司法》有哪些重大变化？
示例回答：
Key Points:
1. 修订强化了对公司治理的监管，明确了董事会和监事会的具体职责。[12-1]
2. 引入了强制性的ESG报告披露要求。[14-2]
3. 调整了公司资本制度，降低了最低注册资本要求。[12-1]
4. 为中小企业引入了专项扶持措施。[15-1][13-2]
Answer:
2023年修订的《公司法》引入了几项重大变化。首先，此次修订强化了公司治理的监管，具体明确了董事会和监事会的职责。[12-1]其次，引入了强制性的环境、社会及公司治理（ESG）报告披露要求。[14-3]此外，修订还调整了公司资本制度，降低了最低注册资本要求。[12-2]最后，此次修订为中小企业引入了专项扶持措施，以促进其发展。[14-2][12-1]
\end{verbatim}
\end{minipage}
\end{CJK*}

\subsection{Error-types Co-occurrence Analysis}\label{app:co-occurences}

\begin{figure*}[t]
    \includegraphics[width=0.49\linewidth]{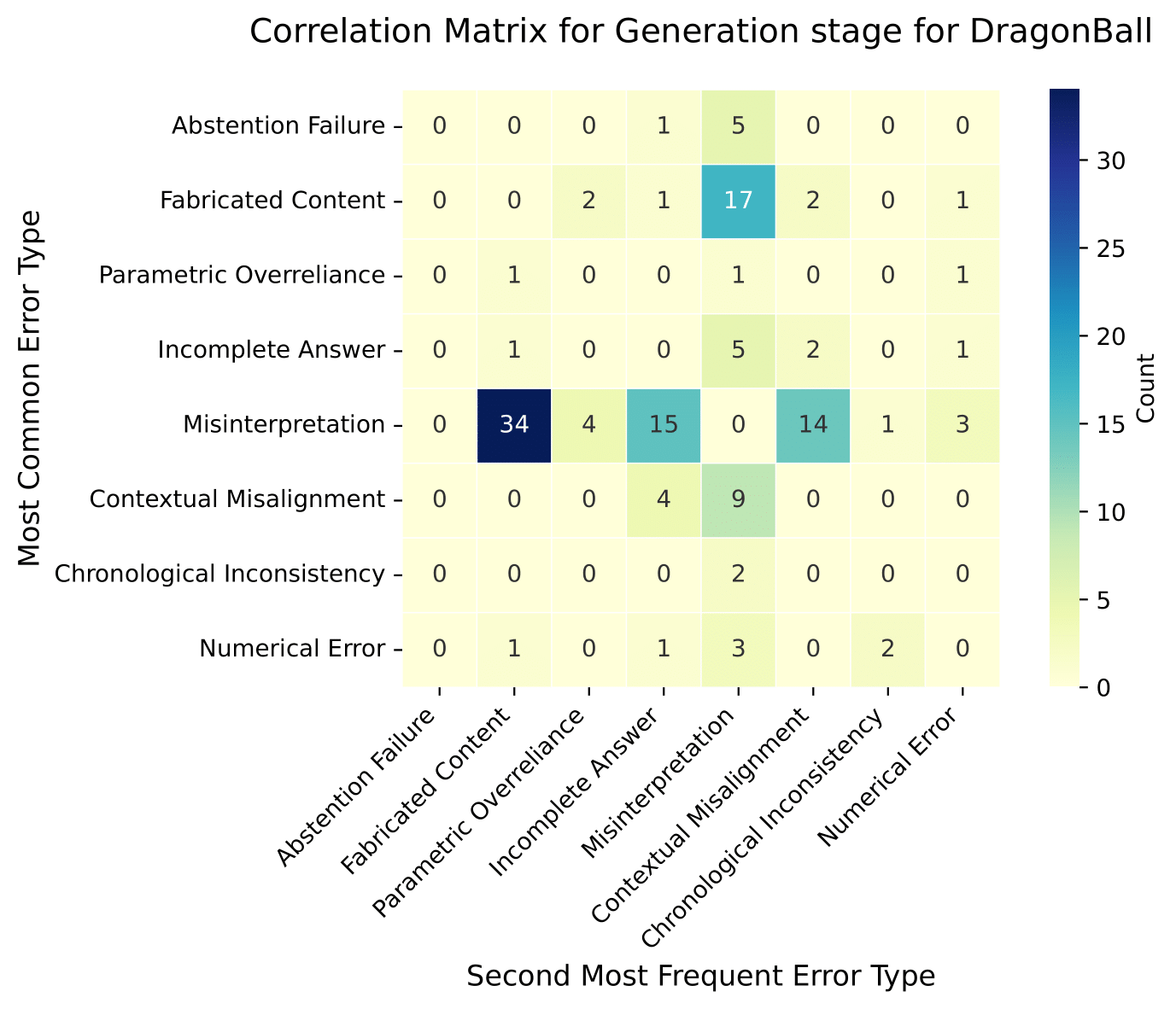}
    \includegraphics[width=0.49\linewidth]{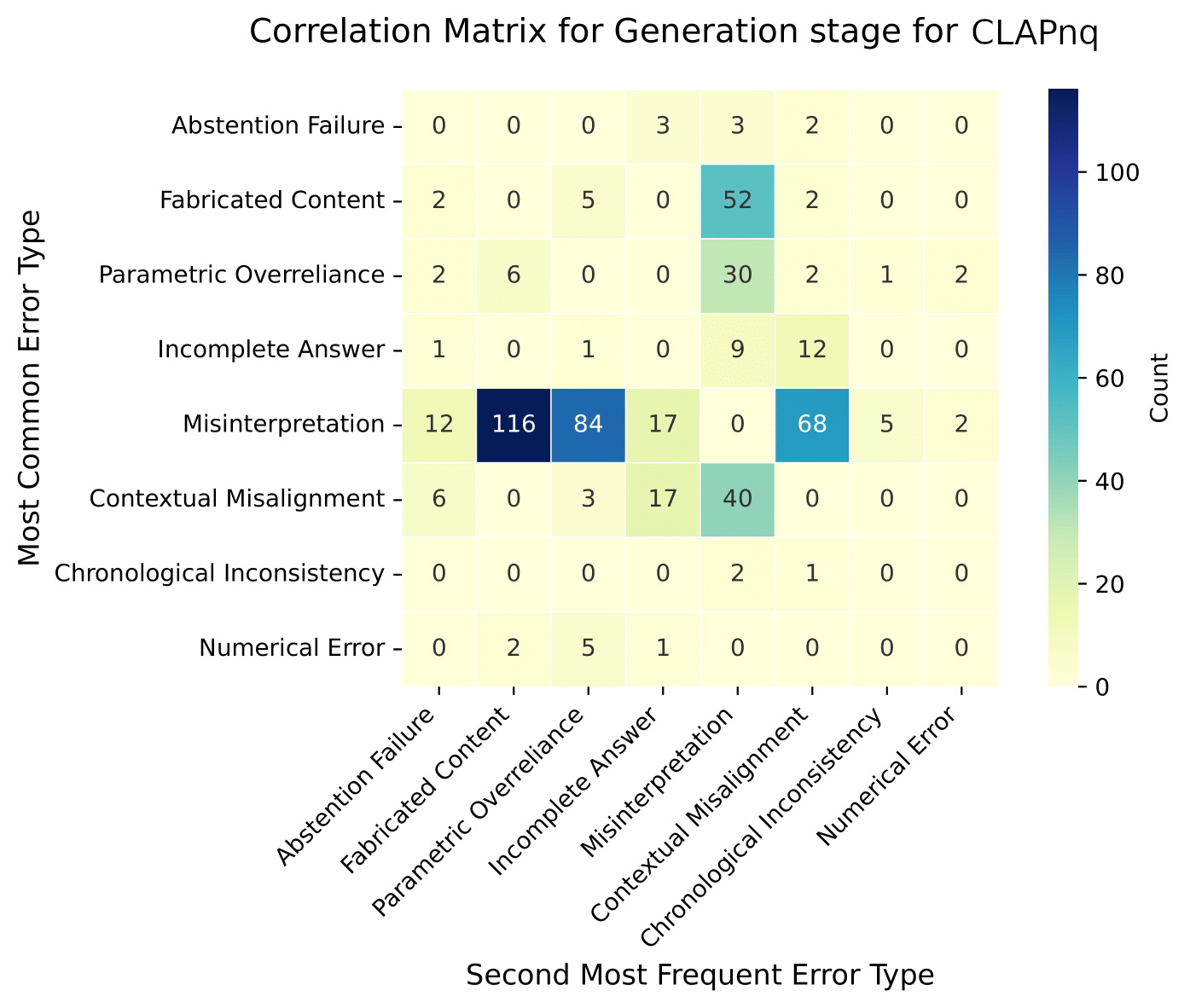}
    \caption{Joint distribution of first and second mode error categories for the generation stage. \textbf{Left:} Dragonball-EN \textbf{Right:} CLAPnq}
    \label{fig:mode-correlation-generation}
\end{figure*}

To understand whether certain error types tend to appear together, we analyze the relationship between the mode and second most frequent error types for each example for each of the RAG stages. 

Figure~\ref{fig:mode-correlation-generation} shows the co-occurrence counts between generation error types for DragonBall-EN and CLAPnq. We observe that \emph{Misinterpretation} \hyperref[e:13]{(E13)} often co-occurs with \emph{Fabricated Content} \hyperref[e:10]{(E10)} and \emph{Contextual Misalignment} \hyperref[e:14]{(E14)}. Interestingly, for CLAPnq, \emph{Misinterpretation} \hyperref[e:13]{(E13)} also frequently co-occurs with \emph{Parameter Overreliance} \hyperref[e:11]{(E11)}, whereas this pattern is not observed in DragonBall-EN. This can be explained by the fact that CLAPnq queries are natural questions that may exist in the model’s parametric knowledge acquired during pre-training, while DragonBall contains domain-specific queries that are unlikely to appear in the pre-training data.

Despite both datasets exhibiting significant $\chi^2$ test results, we observe a notable difference in the density of their error co-occurrence patterns. The contingency table for the DragonBall-EN dataset is considerably sparser, indicating that error types tend to co-occur in fewer, more specific combinations. In contrast, the CLAPnq dataset presents a denser co-occurrence matrix, with a wider variety of error type pairs appearing more frequently. This difference may reflect that CLAPnq surfaces more complex and intertwined error patterns, reinforcing the need for fine-grained diagnostic evaluation across diverse RAG applications.

\subsection{Detailed Error Classification Results}
\begin{figure*}
    \includegraphics[width=\linewidth]{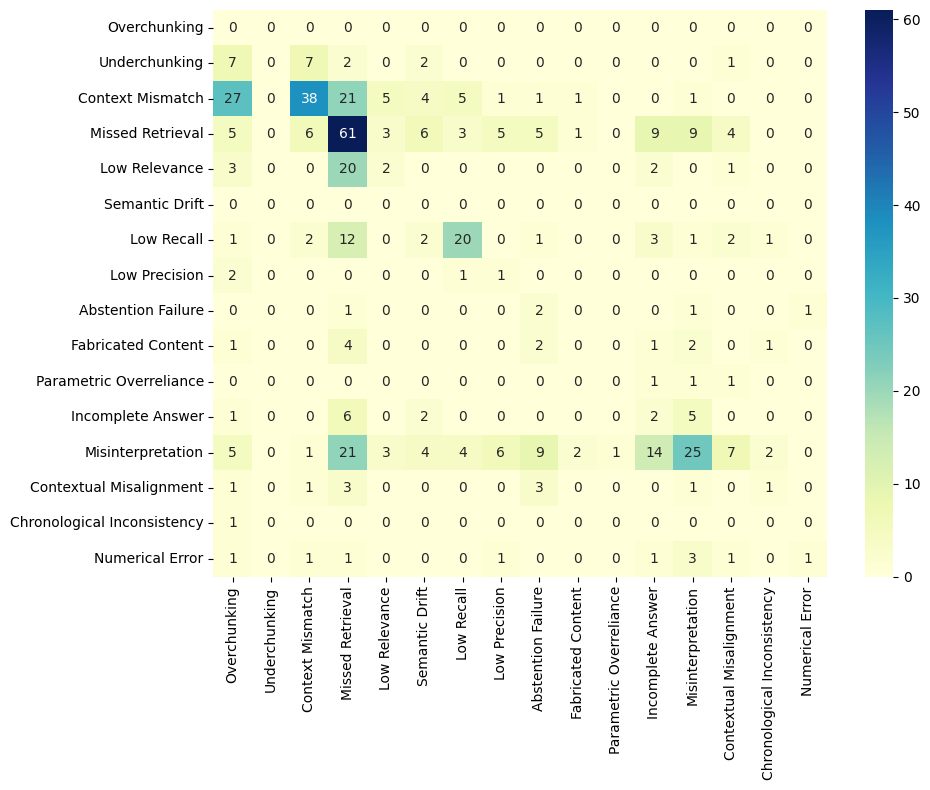}
    \caption{Joint distribution of RAGEC auto-evaluation error type classifications and human annotations for DragonBall-EN}
    \label{fig:confusion}
\end{figure*}

Detailed error classification results comparing RAGEC to human annotations are shown in \Cref{fig:confusion}. For RAGEC, we show the most frequent error type out of the $K=10$ repeated LLM calls per query. Note that humans sometimes annotated more than one error type for a single query (though only from a single stage), and did not indicate an ordering between types. In such cases, each human annotated error type appears in the counts in the figure. We find that some specific error types are preferred by humans or RAGEC, such as \emph{Missed Retrieval} \hyperref[e:4]{(E4)} which was heavily used by humans, or \emph{Context Mismatch} \hyperref[e:3]{(E3)} which was overrepresented in the RAGEC labels.

\section{AI assistant usage}
\label{app:d}
In this research, we used ChatGPT to assist paper editing, literature search, and Github Copilot for coding assistance.

\end{document}